\pdfoutput=1

\documentclass[11pt]{article}

\usepackage[preprint]{acl}
\usepackage{multirow}
\usepackage{adjustbox}
\usepackage{booktabs}

\usepackage[T1]{fontenc}

\usepackage[utf8]{inputenc}

\usepackage{microtype}

\usepackage{inconsolata}
\usepackage{times}
\usepackage{latexsym}
\usepackage{amsmath}
\usepackage{amssymb}
\usepackage{booktabs}
\usepackage{tabularx}

\usepackage{graphicx}

\definecolor{mygrey}{rgb}{0.95,0.95,0.95}
\definecolor{myblue}{rgb}{0.9,0.9,1.0}
\definecolor{myyellow}{rgb}{1.0,1.0,0.9}
\PassOptionsToPackage{table}{xcolor}
\usepackage[table]{xcolor}
\usepackage{colortbl}

\title{Do BERT-Like Bidirectional Models Still Perform Better on Text Classification in the Era of LLMs?}

\author{Junyan Zhang\textsuperscript{\rm 1,*}, Yiming Huang\textsuperscript{\rm 1,*}, Shuliang Liu\textsuperscript{\rm 1,\rm 2},\\
\textbf{Yubo Gao}\textsuperscript{\rm 1},   
\textbf{Xuming Hu}\textsuperscript{\rm 1,\rm 2,}\footnotemark[2]\\
\\
\textsuperscript{\rm 1}The Hong Kong University of Science and Technology (Guangzhou)\\
\textsuperscript{\rm 2}The Hong Kong University of Science and Technology \\
\texttt{junyanzhang0317@gmail.com}, \texttt{huangyiming2002@126.com},
\texttt{xuminghu@hkust-gz.edu.cn}
}

\begin{document}
\maketitle
\renewcommand{\thefootnote}{\fnsymbol{footnote}}
\footnotetext[1]{Equal contribution.}
\footnotetext[2]{Corresponding author.}
\renewcommand{\thefootnote}{\arabic{footnote}}
\begin{abstract}
The rapid adoption of LLMs has overshadowed the potential advantages of traditional BERT-like models in text classification. This study challenges the prevailing ``LLM-centric'' trend by systematically comparing three category methods, \textit{i.e.,} BERT-like models fine-tuning, LLM internal state utilization, and zero-shot inference across six high-difficulty datasets. Our findings reveal that BERT-like models often outperform LLMs. We further categorize datasets into three types, perform PCA and probing experiments, and identify task-specific model strengths: BERT-like models excel in pattern-driven tasks, while LLMs dominate those requiring deep semantics or world knowledge. Based on this, we propose \textbf{TaMAS}, a fine-grained task selection strategy, advocating for a nuanced, task-driven approach over a one-size-fits-all reliance on LLMs.
\end{abstract}

\section{Introduction and Related Work}

With the rise of Large Language Models (LLMs), the text classification research paradigm is shifting significantly. The academic community currently exhibits a pronounced ``LLM-centric'' trend \cite{li2024llms, xie2024large}, \textit{i.e.,} an increasing number of studies focus on enhancing the classification performance of LLMs through techniques such as prompt engineering \cite{xiao2024toxicloakcn, zhang2023interpretable}, internal state extraction \cite{marks2023geometry, azaria2023internal}, or parameter-efficient fine-tuning \cite{inan2023llama, zhang2024pushing}. However, this trend overlooks a critical issue: traditional BERT-like models \cite{devlin2019bert, liu2019roberta} may still hold unique advantages in certain key scenarios. Notably, even SOTA LLMs achieve only marginal and costly performance gains on challenging tasks like implicit hate speech detection involving homophones or emoji substitutions \cite{xiao2024toxicloakcn}.

\begin{figure}[ht]
    \centering
    \includegraphics[width=0.46\textwidth]{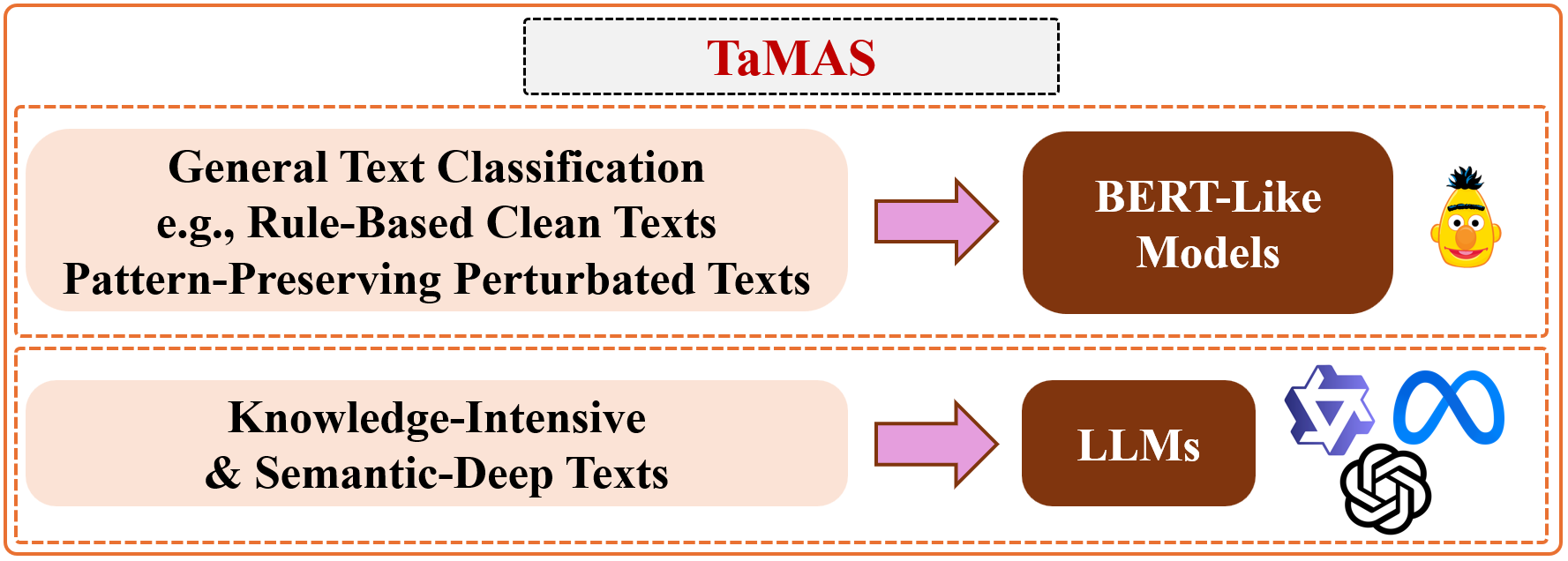} 
    \caption{Illustration of our fine-grained task selection strategy TaMAS.}
    \label{fig:framework}
\end{figure}

Unlike previous studies that focused mainly on LLM applications in single-type text classification \cite{zhang2024pushing} or evaluated text classification tasks without sufficient interpretability analysis or categorization \cite{vajjala2025text}, this work identifies an overlooked research gap and raises a key question: \textit{Under the LLM-dominated paradigm, have we prematurely overlooked the potential of BERT-like models?} To address this question, we performed a comprehensive comparative evaluation in the field of text classification. For the first time, we systematically examined the performance boundaries of three computationally low-cost mainstream methods in the LLM era, i.e., BERT-like models fine-tuning, LLM internal state utilization, and LLM zero-shot inference, across six high-difficulty tasks.

Empirical results have yielded a groundbreaking finding: For the majority of classification tasks, BERT-like models are more suitable, as they require fewer computational resources while maintaining high performance. In addition, different methods demonstrate varying performance across different datasets.

To better understand this discrepancy, we conduct a comprehensive analysis through PCA visualization of the model internals and perform probing experiments from interpretable perspectives. Our findings indicate that datasets can be categorized into three main types, offering insights into the factors driving model performance differences. Based on this, we establish \textbf{TaMAS}, a fine-grained \textbf{T}ask-\textbf{a}ware \textbf{M}odel \textbf{A}daptation \textbf{S}trategy shown in Figure \ref{fig:framework}, which reveals: For basic text classification tasks with discernible textual patterns, even after perturbations, BERT-like models consistently outperform LLM-based approaches. For tasks demanding deep semantic understanding or real-world knowledge (\textit{e.g.,} complex reasoning is needed or hallucination detection), LLMs hold a clear advantage. These findings not only provide a scientific basis for model selection but also critique the prevailing ``LLM-first'' trend in research. 

In summary, our contribution can be concluded in the following threefold: (1) We rigorously and comprehensively reaffirmed the technical standing of BERT-like models in text classification tasks through extensive experimentation. (2) Based on performance across six datasets, we classify them into three types. Through in-depth model internal analysis and visualization, we explore why different methods excel in each case. (3) We proposed \textbf{TaMAS} for classification methodologies.

\begin{table*}[t]
\centering
\resizebox{1\textwidth}{!}
{
\renewcommand{\arraystretch}{1.1}
\begin{tabular}{ll*{18}{c}}
\toprule
\multirow{2}{*}{\textbf{Cat.}} & \textbf{Datasets}
& \multicolumn{3}{c}{\textbf{ToxiCloakCNBase}}
& \multicolumn{3}{c}{\textbf{ToxiCloakCNEmoji}}
& \multicolumn{3}{c}{\textbf{ToxiCloakCNHomo}}
& \multicolumn{3}{c}{\textbf{LegalText}}
& \multicolumn{3}{c}{\textbf{MaliciousCode}}
& \multicolumn{3}{c}{\textbf{Hallucination}} \\
\cmidrule(lr){3-5} \cmidrule(lr){6-8} \cmidrule(lr){9-11} \cmidrule(lr){12-14} \cmidrule(lr){15-17} \cmidrule(lr){18-20}
& \textbf{Meth./Met.} & \multicolumn{1}{c}{\textbf{AUC}} & \multicolumn{1}{c}{\textbf{Acc}} & \multicolumn{1}{c}{\textbf{F1}}
& \multicolumn{1}{c}{\textbf{AUC}} & \multicolumn{1}{c}{\textbf{Acc}} & \multicolumn{1}{c}{\textbf{F1}}
& \multicolumn{1}{c}{\textbf{AUC}} & \multicolumn{1}{c}{\textbf{Acc}} & \multicolumn{1}{c}{\textbf{F1}}
& \multicolumn{1}{c}{\textbf{AUC}} & \multicolumn{1}{c}{\textbf{Acc}} & \multicolumn{1}{c}{\textbf{F1}}
& \multicolumn{1}{c}{\textbf{AUC}} & \multicolumn{1}{c}{\textbf{Acc}} & \multicolumn{1}{c}{\textbf{F1}}
& \multicolumn{1}{c}{\textbf{AUC}} & \multicolumn{1}{c}{\textbf{Acc}} & \multicolumn{1}{c}{\textbf{F1}} \\
\cmidrule(lr){1-20} 

\rowcolor{mygrey}
& BERT
& 95.6 & 88.1 & 88.3 & 92.0 & 85.4 & 84.9 & 91.2 & 82.8 & 82.3
& 98.4 & 93.3 & 93.4 & 99.7 & 99.7 & 99.7 & 76.6 & 65.2 & 63.9 \\
\rowcolor{mygrey}
& RoBERTa
& 95.5 & 88.7 & 88.4 & 91.0 & 83.5 & 83.3 & 90.6 & 81.5 & 82.4
& 99.2 & 96.0 & 96.0 & 99.9 & 99.3 & 99.3 & 84.2 & 72.7 & 75.6 \\
\rowcolor{mygrey}
& ERNIE
& 96.0 & 89.6 & 89.4 & 92.2 & 83.3 & 83.6 & 91.8 & 84.5 & 84.7
& 98.5 & 93.7 & 93.7 & 99.7 & 99.7 & 99.7 & 81.6 & 71.5 & 72.2 \\
\rowcolor{mygrey}
\multirow{-4}{*}{BLMs} & ELECTRA
& 95.1 & 87.4 & 87.0 & 89.4 & 80.9 & 80.9 & 88.9 & 81.0 & 81.3
& 98.7 & 93.3 & 93.3 & 99.7 & 99.7 & 99.7 & 85.5 & 75.7 & 75.8 \\
\cmidrule(lr){1-20}

\rowcolor{myyellow}
& SAPLMA\textsubscript{prism}
& 92.7 & 83.2 & 82.1 & 87.1 & 79.0 & 78.5 & 84.4 & 75.5 & 76.4
& 97.7 & 92.3 & 92.4 & 100.0 & 99.7 & 99.7 & 95.9 & 89.3 & 90.0 \\
\rowcolor{myyellow}
\multirow{-2}{*}{LLM-IS} & MM-Probe\textsubscript{prism}
& 88.2 & 78.3 & 76.6 & 83.3 & 75.7 & 74.9 & 80.9 & 72.6 & 69.4
& 91.3 & 83.3 & 84.5 & 100.0 & 98.9 & 99.0 & 93.5 & 86.1 & 86.1  \\

\cmidrule(lr){1-20}

\rowcolor{myblue}
& Query\textsubscript{Qwen/LLaMA}
& 72.8 & 72.8 & 65.9 & 69.1 & 69.1 & 61.5 & 68.6 & 68.6 & 59.3
& 80.3 & 80.3 & 81.7 & 96.1 & 96.1 & 96.3 & 85.7 & 85.7 & 86.2 \\
\rowcolor{myblue}
\multirow{-2}{*}{LLM-Q} & Query\textsubscript{GPT-4o}
& - & - & 79.6 & - & - & 75.4 & - & - & 74.1
& - & - & - & - & - & - & - & - & - \\
\bottomrule
\end{tabular}
}
\caption{Evaluation of different methods on six datasets using AUC, accuracy, and F1 score. Meth. and Met. stand for Methods and Metrics. BLMs refers to BERT-like models, while LLM-IS and LLM-Q denote approaches using LLM internal states and direct querying, respectively. Cat. indicates Categories.}
\label{tab:main_results_merged}
\end{table*}

\section{Comprehensive Test Across Six Typical Datasets}

In this section, we conduct comprehensive experiments across six typical datasets and three major categories of text classification methods.

\subsection{Experimental Setup}

\paragraph{Compared Methods:}

For BERT-like models, we selected four variants: BERT \cite{devlin2019bert}, RoBERTa \cite{liu2019roberta}, ERNIE \cite{sun2020ernie, sun2021ernie}, and ELECTRA \cite{clark2020electra}. LLM-based methods include two types of methods. For methods leveraging the internal states of LLMs, we selected SAPLMA \citep{azaria2023internal} and MM-Probe \cite{marks2023geometry}. For both of these methods, we applied the Prism \cite{zhang2024prompt} approach to enhance their performance. For LLM zero-shot querying method, we ask LLMs to output results directly. Additionally, for the ToxiCloakCN dataset, we incorporate the best results reported by \citet{xiao2024toxicloakcn}. For more information about model and implementation details, please refer to \S\ref{sec:Model Details}, \S\ref{sec:Implementation Details}, and \S\ref{sec:LLM-IS Implementations}.

\paragraph{Evaluation Metrics:}

We used classic metrics for evaluating binary classification tasks, including AUC, Accuracy (Acc), and F1 score, in order to comprehensively assess model performance. 

\paragraph{Datasets:}
   
We selected six representative datasets that are moderately challenging and feature a degree of novelty. Specifically, we selected the ToxiCloakCN \cite{xiao2024toxicloakcn}, Hallucination \cite{azaria2023internal}, MaliciousCode \cite{Er1111c_Malicious_code_classification}, and LegalText \cite{opensuse_cavil_legal_text} datasets. These datasets are used for detecting implicit hate speech, hallucinations, malicious code, and legal text in source code, respectively. Specifically, ToxiCloakCN consists of three parts of chinese hate speech data, including the base data, data perturbed with homophone substitution, and data perturbed with emoji substitution. For more information about datasets details, please refer to \S\ref{sec:Datasets Details}.

\begin{figure*}[ht]
    \centering 
    \includegraphics[width=0.87\textwidth]{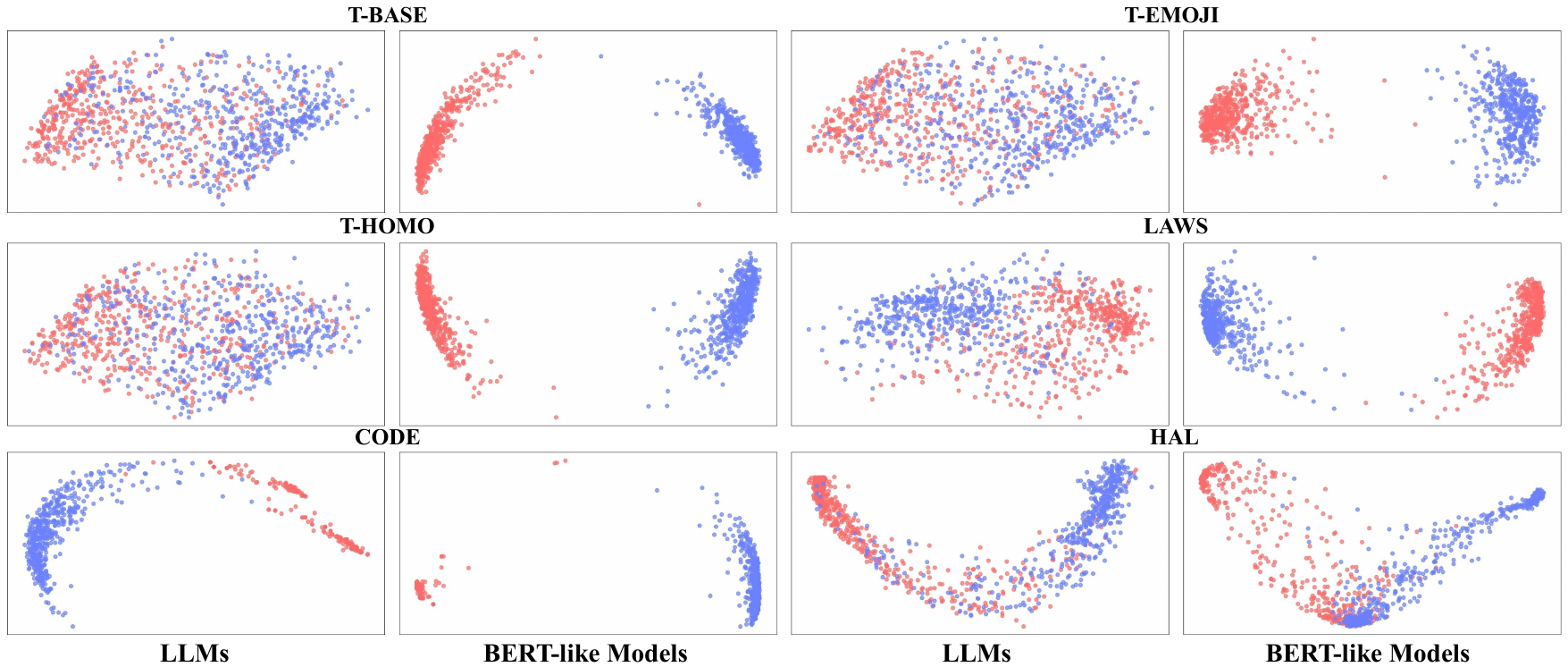}
    \caption{Comparative PCA visualization of hidden states across six datasets: BERT-like models vs. LLMs. T-BASE, T-EMOJI, T-HOMO, LAWS, CODE, HAL refer to ToxiCloakCNBase, ToxiCloakCNEmoji, ToxiCloakCNHomo, LegalText, MaliciousCode, Hallucination datasets.}
    \label{fig:pca}
\end{figure*}

\begin{figure}[ht]
    \centering 
    \includegraphics[width=0.46\textwidth]{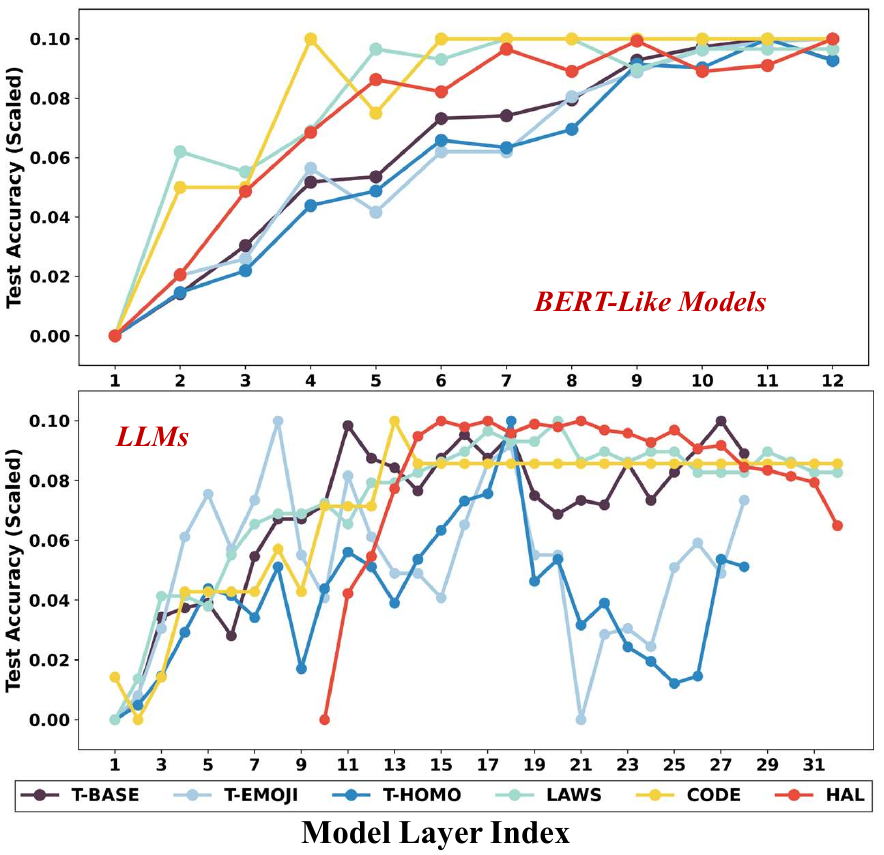} 
    \caption{Comparative visualization of hidden states classification separability using single linear probes on all datasets: BERT-like models vs. LLMs. The fundamental difference in how BERT-like models and LLMs process information becomes particularly evident in the layerwise progression of separability.}
    \label{fig:probe} 
\end{figure}

\subsection{Results}

Based on the experimental results shown in Table \ref{tab:main_results_merged}, we can conclude that in the era of LLMs, BERT-like models still demonstrate strong performance on a wide range of text classification tasks. Additionally, on different datasets, BERT-like models and LLM-based methods demonstrate varying performance. For example, on the ToxiCloakCNHomo dataset, BERT-like models exhibit outstanding performance, whereas on hallucination datasets, they underperform compared to methods utilizing LLM internal states and LLM zero-shot inference. 

\section{Analysis \& Discussions}

Based on experimental results, this chapter classifies six datasets into \textit{\textbf{three}} categories and analyzes performance using model hidden states PCA visualization and probing.

Performance rankings differ by dataset type: For three implicit hate speech datasets, BERT-like methods are superior, followed by those using LLM internal states, with direct LLM querying performing poorest. Malicious code and legal text detection show BERT-like and LLM internal state methods performing competitively and better than direct querying. For hallucination detection, LLM internal state methods surpass direct querying, both exceeding BERT-like model performances.

Motivated by the observed performance differences, we conducted the computation and visualization of Figures \ref{fig:pca} and \ref{fig:probe}, and then performed an in-depth analysis of the characteristics of these dataset types to better understand the sources of variation. For more information of the details of these figures, please refer to \S\ref{sec:Visualization Details}.

\paragraph{Pattern-Preserving Perturbated Texts} 

Implicit hate speech datasets feature substantial covert language, from basic euphemisms (ToxiCloakCNBase) to more sophisticated emoji and homophone substitutions (ToxiCloakCNEmoji, ToxiCloakCNHomo). This linguistic obfuscation increases semantic opacity and requires specific contextual knowledge, which is often unavailable to non-community members, including LLMs.

From a modeling perspective, directly querying LLMs via prompting strategies for classification purposes yields suboptimal results. This limitation primarily stems from the fact that such models have had limited exposure to these highly concealed linguistic patterns during their pre-training phase. Moreover, even when employing methods based on internal representations, \textit{i.e.,} such as probing techniques, the classification performance remains constrained. At the same time, the presence of covert language, which, under the conventional pre-training objective of next-token prediction, introduces a significant amount of redundant or misleading information into the internal representations of LLMs. As Figure \ref{fig:pca} shows, LLM hidden layer representations of perturbed and implicit hate speech are poorly separated and intermingled.

Notably, although the three categories of subtle hate speech are semantically challenging to distinguish, close observation reveals a degree of coherence and regularity in the use of covert expressions. For example, there exist systematic patterns in the deployment of emojis and structural consistencies in homophonic substitutions. These linguistic phenomena fundamentally rely on contextual understanding, an area in which bidirectional attention-based models like BERT demonstrate particular strength. As depicted in Figure \ref{fig:pca}, the \texttt{CLS} token embeddings demonstrate a high degree of linear separability between harmful and harmless instances. Consequently, such models exhibit superior performance on this type of task.

\paragraph{Rule-Based Clean Texts} 

In malicious code and legal text datasets, both BERT-like models and methods using LLM internal states perform competitively. While challenging for non-experts, expert analysis shows the discriminative patterns are coherent, regular, and rule-based. Crucially, unlike heavily obfuscated data from previous tasks, the current data is largely clean, free from perturbations or complex euphemisms. This data is well-represented in LLM pre-training, facilitating high-quality representations. As Figure \ref{fig:pca} illustrates, PCA on BERT and LLM hidden representations reveals relatively good class separability, supporting the strong performance of methods like SAPLMA and MM-Probe, comparable to BERT-based models. Directly querying LLMs shows weaker performance than BERT-based or LLM representation-based methods, likely due to misalignment between LLMs and human decision-making criteria \cite{jiang2023evaluating}.

\paragraph{Knowledge-Intensive \& Semantic-Deep Texts}

For hallucination detection, methods leveraging LLM internal states outperform other approaches.

This task is distinct as it requires not just natural language understanding but crucially, comparison with real-world knowledge to assess truthfulness. LLMs, due to their scale and extensive pretraining data, acquire vast real-world knowledge. Research indicates LLMs develop internal directions representing abstract concepts \cite{arditi2024refusal}, including truthfulness \cite{marks2023geometry, azaria2023internal}, suggesting their internal representations are inherently better suited for capturing truth. In contrast, BERT-like models struggle because hallucination detection datasets are limited relative to the breadth of real-world facts, hindering their ability to learn reliable representations for concepts like “truthfulness”.

Furthermore, influenced by the next-token prediction objective, when an LLM detects a contradiction in the encoded input, it may encode signals of untruth in the final token’s hidden state, anticipating generating tokens like ``false'' or attempting to correct its own error, shown in Table \ref{tab:Generation Examples}. This aligns with the findings proposed by \citet{azaria2023internal}. This results in stronger distinguishability between hallucinated and non-hallucinated statements at the hidden state level, providing a theoretical basis for state-based detection methods.

Consequently, methods utilizing LLM internal representations achieve superior performance in hallucination detection tasks.

\section{TaMAS}

Based on our findings, we propose \textbf{TaMAS}, a fine-grained strategy shown in Figure \ref{fig:framework} which guides the selection of BERT-like models or LLMs based on the characteristics of the texts.

For \textbf{General Text Classification} tasks that primarily rely on surface-form patterns or shallow semantic features, BERT-like models demonstrate superior parameter efficiency and performance. 

In contrast, there are two critical scenarios where conventional BERT-like models exhibit limitations:

\textbf{Knowledge-Intensive Classification:} When the task requires substantial domain-specific prerequisite knowledge that cannot be adequately covered by existing training datasets. This typically occurs where the label determination depends on implicit knowledge beyond surface-level textual patterns. 

\textbf{Semantic-Deep Classification:} Cases where accurate categorization demands profound semantic understanding that cannot be reliably inferred from lexical features alone.

\section{Conclusion}

Our study challenges the prevailing ``LLM-centric'' trend in text classification by demonstrating that BERT-like models often outperform LLMs while being computationally efficient. Through extensive experiments, we identify three dataset types and propos \textbf{TaMAS}, a fine-grained strategy guiding optimal model choice based on task requirements. This work advocates for a rational, task-driven approach over blind adherence to LLMs, ensuring efficiency without sacrificing performance. 

\clearpage

\section*{Limitations}

This paper mainly explores six typical and challenging datasets, and focuses on investigating three major categories of methods. Our future work aims to conduct experiments on a broader range of datasets and evaluate them using a wider variety of approaches, in order to draw more comprehensive conclusions and develop effective task-specific selection strategies.

\bibliography{acl_latex}

\begin{thebibliography}{22}
\providecommand{\natexlab}[1]{#1}

\bibitem[{Alain and Bengio(2016)}]{alain2016understanding}
Guillaume Alain and Yoshua Bengio. 2016.
\newblock Understanding intermediate layers using linear classifier probes.
\newblock \emph{arXiv preprint arXiv:1610.01644}.

\bibitem[{Arditi et~al.(2024)Arditi, Obeso, Syed, Paleka, Panickssery, Gurnee, and Nanda}]{arditi2024refusal}
Andy Arditi, Oscar Obeso, Aaquib Syed, Daniel Paleka, Nina Panickssery, Wes Gurnee, and Neel Nanda. 2024.
\newblock Refusal in language models is mediated by a single direction.
\newblock \emph{arXiv preprint arXiv:2406.11717}.

\bibitem[{Azaria and Mitchell(2023)}]{azaria2023internal}
Amos Azaria and Tom Mitchell. 2023.
\newblock The internal state of an llm knows when it's lying.
\newblock \emph{arXiv preprint arXiv:2304.13734}.

\bibitem[{Chen et~al.(2024)Chen, Liu, Chen, Gu, Wu, Tao, Fu, and Ye}]{chen2024inside}
Chao Chen, Kai Liu, Ze~Chen, Yi~Gu, Yue Wu, Mingyuan Tao, Zhihang Fu, and Jieping Ye. 2024.
\newblock Inside: Llms' internal states retain the power of hallucination detection.
\newblock \emph{arXiv preprint arXiv:2402.03744}.

\bibitem[{Clark et~al.(2020)Clark, Luong, Le, and Manning}]{clark2020electra}
Kevin Clark, Minh-Thang Luong, Quoc~V Le, and Christopher~D Manning. 2020.
\newblock Electra: Pre-training text encoders as discriminators rather than generators.
\newblock \emph{arXiv preprint arXiv:2003.10555}.

\bibitem[{Devlin et~al.(2019)Devlin, Chang, Lee, and Toutanova}]{devlin2019bert}
Jacob Devlin, Ming-Wei Chang, Kenton Lee, and Kristina Toutanova. 2019.
\newblock Bert: Pre-training of deep bidirectional transformers for language understanding.
\newblock In \emph{Proceedings of the 2019 conference of the North American chapter of the association for computational linguistics: human language technologies, volume 1 (long and short papers)}, pages 4171--4186.

\bibitem[{Er1111c(2024)}]{Er1111c_Malicious_code_classification}
Er1111c. 2024.
\newblock Malicious code classification.
\newblock \url{https://huggingface.co/datasets/Er1111c/Malicious_code_classification}.

\bibitem[{Inan et~al.(2023)Inan, Upasani, Chi, Rungta, Iyer, Mao, Tontchev, Hu, Fuller, Testuggine et~al.}]{inan2023llama}
Hakan Inan, Kartikeya Upasani, Jianfeng Chi, Rashi Rungta, Krithika Iyer, Yuning Mao, Michael Tontchev, Qing Hu, Brian Fuller, Davide Testuggine, and 1 others. 2023.
\newblock Llama guard: Llm-based input-output safeguard for human-ai conversations.
\newblock \emph{arXiv preprint arXiv:2312.06674}.

\bibitem[{Jiang et~al.(2023)Jiang, Xu, Zhu, Han, Zhang, and Zhu}]{jiang2023evaluating}
Guangyuan Jiang, Manjie Xu, Song-Chun Zhu, Wenjuan Han, Chi Zhang, and Yixin Zhu. 2023.
\newblock Evaluating and inducing personality in pre-trained language models.
\newblock \emph{Advances in Neural Information Processing Systems}, 36:10622--10643.

\bibitem[{Li et~al.(2024)Li, Dong, Chen, Su, Zhou, Ai, Ye, and Liu}]{li2024llms}
Haitao Li, Qian Dong, Junjie Chen, Huixue Su, Yujia Zhou, Qingyao Ai, Ziyi Ye, and Yiqun Liu. 2024.
\newblock Llms-as-judges: a comprehensive survey on llm-based evaluation methods.
\newblock \emph{arXiv preprint arXiv:2412.05579}.

\bibitem[{Liu et~al.(2019)Liu, Ott, Goyal, Du, Joshi, Chen, Levy, Lewis, Zettlemoyer, and Stoyanov}]{liu2019roberta}
Yinhan Liu, Myle Ott, Naman Goyal, Jingfei Du, Mandar Joshi, Danqi Chen, Omer Levy, Mike Lewis, Luke Zettlemoyer, and Veselin Stoyanov. 2019.
\newblock Roberta: A robustly optimized bert pretraining approach.
\newblock \emph{arXiv preprint arXiv:1907.11692}.

\bibitem[{Marks and Tegmark(2023)}]{marks2023geometry}
Samuel Marks and Max Tegmark. 2023.
\newblock The geometry of truth: Emergent linear structure in large language model representations of true/false datasets.
\newblock \emph{arXiv preprint arXiv:2310.06824}.

\bibitem[{openSUSE(2025)}]{opensuse_cavil_legal_text}
openSUSE. 2025.
\newblock Cavil legal text dataset.
\newblock \url{https://huggingface.co/datasets/openSUSE/cavil-legal-text}.

\bibitem[{Skean et~al.(2024)Skean, Arefin, LeCun, and Shwartz-Ziv}]{skean2024does}
Oscar Skean, Md~Rifat Arefin, Yann LeCun, and Ravid Shwartz-Ziv. 2024.
\newblock Does representation matter? exploring intermediate layers in large language models.
\newblock \emph{arXiv preprint arXiv:2412.09563}.

\bibitem[{Sun et~al.(2021)Sun, Wang, Feng, Ding, Pang, Shang, Liu, Chen, Zhao, Lu et~al.}]{sun2021ernie}
Yu~Sun, Shuohuan Wang, Shikun Feng, Siyu Ding, Chao Pang, Junyuan Shang, Jiaxiang Liu, Xuyi Chen, Yanbin Zhao, Yuxiang Lu, and 1 others. 2021.
\newblock Ernie 3.0: Large-scale knowledge enhanced pre-training for language understanding and generation.
\newblock \emph{arXiv preprint arXiv:2107.02137}.

\bibitem[{Sun et~al.(2020)Sun, Wang, Li, Feng, Tian, Wu, and Wang}]{sun2020ernie}
Yu~Sun, Shuohuan Wang, Yukun Li, Shikun Feng, Hao Tian, Hua Wu, and Haifeng Wang. 2020.
\newblock Ernie 2.0: A continual pre-training framework for language understanding.
\newblock In \emph{Proceedings of the AAAI conference on artificial intelligence}, volume~34, pages 8968--8975.

\bibitem[{Vajjala and Shimangaud(2025)}]{vajjala2025text}
Sowmya Vajjala and Shwetali Shimangaud. 2025.
\newblock Text classification in the llm era--where do we stand?
\newblock \emph{arXiv preprint arXiv:2502.11830}.

\bibitem[{Xiao et~al.(2024)Xiao, Hu, Choo, and Lee}]{xiao2024toxicloakcn}
Yunze Xiao, Yujia Hu, Kenny Tsu~Wei Choo, and Roy Ka-wei Lee. 2024.
\newblock Toxicloakcn: Evaluating robustness of offensive language detection in chinese with cloaking perturbations.
\newblock \emph{arXiv preprint arXiv:2406.12223}.

\bibitem[{Xie et~al.(2024)Xie, Chen, Zhang, Wan, and Li}]{xie2024large}
Junlin Xie, Zhihong Chen, Ruifei Zhang, Xiang Wan, and Guanbin Li. 2024.
\newblock Large multimodal agents: A survey.
\newblock \emph{arXiv preprint arXiv:2402.15116}.

\bibitem[{Zhang et~al.(2024{\natexlab{a}})Zhang, Yu, Yi, Zhang, Li, and Liu}]{zhang2024prompt}
Fujie Zhang, Peiqi Yu, Biao Yi, Baolei Zhang, Tong Li, and Zheli Liu. 2024{\natexlab{a}}.
\newblock Prompt-guided internal states for hallucination detection of large language models.
\newblock \emph{arXiv preprint arXiv:2411.04847}.

\bibitem[{Zhang et~al.(2023)Zhang, Luo, Chuang, Fang, Gaitskell, Hartvigsen, Wu, Fox, Meng, and Glass}]{zhang2023interpretable}
Tianhua Zhang, Hongyin Luo, Yung-Sung Chuang, Wei Fang, Luc Gaitskell, Thomas Hartvigsen, Xixin Wu, Danny Fox, Helen Meng, and James Glass. 2023.
\newblock Interpretable unified language checking.
\newblock \emph{arXiv preprint arXiv:2304.03728}.

\bibitem[{Zhang et~al.(2024{\natexlab{b}})Zhang, Wang, Ren, Li, Tiwari, Wang, and Qin}]{zhang2024pushing}
Yazhou Zhang, Mengyao Wang, Chenyu Ren, Qiuchi Li, Prayag Tiwari, Benyou Wang, and Jing Qin. 2024{\natexlab{b}}.
\newblock Pushing the limit of llm capacity for text classification.
\newblock \emph{arXiv preprint arXiv:2402.07470}.

\end{thebibliography}

\appendix

\section{Model Details}
\label{sec:Model Details}
For the BERT-like models, the specific models selected were as follows: For chinese tasks: bert-base-chinese, roberta-chinese-base, ernie-3.0-base-zh, and chinese-electra-180g-base-discriminator. For english tasks: bert-base-uncased, roberta-base, ernie-2.0-base-en, and electra-base-discriminator. Since there is no version 3.0 available for ERNIE in english, we opted for the 2.0 version instead. The number of parameters in all the models are available in Table \ref{tab:model_parameters}.

\begin{table}[htbp]
  \centering
  \resizebox{\columnwidth}{!}{
  \begin{tabular}{lc}
    \toprule
    \textbf{Model Names} & \textbf{Parameters} \\
    \midrule
    bert-base-chinese & 102.27 M \\
    roberta-chinese-base & 102.27 M \\
    ernie-3.0-base-zh & 117.94 M \\
    chinese-electra-180g-base-discriminator & 101.68 M \\
    bert-base-uncased & 109.48 M \\
    roberta-base & 124.65 M \\
    ernie-2.0-base-en & 109.48 M \\
    electra-base-discriminator & 108.89 M \\
    Qwen2.5-7B-Instruct & 7070.62 M \\
    LLaMA-3-8B-Instruct & 7504.92 M \\
    \bottomrule
  \end{tabular}
  }
  \caption{Model parameter details.}
  \label{tab:model_parameters}
\end{table}

\section{Implementation Details}
\label{sec:Implementation Details}

In the implementation, for each dataset, we split the data into training, validation, and test sets with a ratio of 7:1.5:1.5. For trainable models, we selected the best-performing model on the validation set and evaluated it on the test set. For BERT-like models, we set the learning rate to 2e-5, trained for 10 epochs, and set the dropout rate to 0.5. We do not fine-tune any of these hyper-parameters for this task. For LLM zero-shot querying, we used Qwen2.5-7B-Instruct for chinese text classification tasks and LLaMA-3-8B-Instruct for english text classification tasks. We do not include methods that involve fine-tuning the parameters of the LLMs, as this would significantly increase the computational cost. 

\section{LLM-IS Implementations}
\label{sec:LLM-IS Implementations}

For SAPLMA and MM-Probe, we both choose the 3/4th layer, as middle-to-late layers of the LLM have been proven to potentially better capture the overall sentence semantics \cite{skean2024does, chen2024inside, azaria2023internal}.

\paragraph{SAPLMA.} For the SAPLMA method, which uses the MLP classifier, we set the hidden layer dimensions to [512, 256, 128] all utilizing ReLU activations. The final layer is a sigmoid output. The learning rate is 1e-3. 

\paragraph{MM-Probe.} 
We calculate the mean activation $\mu_i^{(l)}$ for positive examples from $\mathcal{D}_{\text{train}}^{\text{positive}}$ and $\nu_i^{(l)}$ for negative examples from $\mathcal{D}_{\text{train}}^{\text{negative}}$.

\begin{equation}
\begin{aligned}
\mu_i^{(l)} &= \frac{1}{\left| \mathcal{D}_{\text{train}}^{\text{positive}} \right|} 
\sum_{t \in \mathcal{D}_{\text{train}}^{\text{positive}}} x_i^{(l)}(t), \\
\nu_i^{(l)} &= \frac{1}{\left| \mathcal{D}_{\text{train}}^{\text{negative}} \right|} 
\sum_{t \in \mathcal{D}_{\text{train}}^{\text{negative}}} x_i^{(l)}(t).
\end{aligned}
\label{eq:mu_nu}
\end{equation}

We then compute the mass-mean vector for further classification:
\begin{equation}
r_i^{(l)} = \mu_i^{(l)} - \nu_i^{(l)}.
\label{eq:diff_means}
\end{equation}

For the MM-Probe method, we set the classification threshold based on the value that derives the maximum G-Mean, and calculated Acc and F1 scores using this threshold.

\section{Datasets Details}
\label{sec:Datasets Details}

For the ToxiCloakCNBase, ToxiCloakCNBaseEmoji, ToxiCloakCNBaseHomo, and Hallucination datasets, where the number of samples in the two classes is nearly equal, we applied stratified sampling. For the MaliciousCo dataset, we performed undersampling. As for the LegalText dataset, we extracted a balanced subset of two thousand samples. The source information for the datasets used can be found in the Table \ref{tab:dataset link}.

\begin{table*}[htbp]
    \centering
    \resizebox{\textwidth}{!}{
    \begin{tabular}{lc}
        \toprule
        \textbf{Dataset} & \textbf{Link} \\
        \midrule
        ToxiCloakCNBase & \url{https://github.com/Social-AI-Studio/ToxiCloakCN/tree/main} \\
        ToxiCloakCNEmoji & \url{https://github.com/Social-AI-Studio/ToxiCloakCN/tree/main} \\
        ToxiCloakCNHomo & \url{https://github.com/Social-AI-Studio/ToxiCloakCN/tree/main} \\
        LegalText & \url{https://huggingface.co/datasets/openSUSE/cavil-legal-text} \\
        MaliciousCode & \url{https://huggingface.co/datasets/Er1111c/Malicious_code_classification} \\
        Hallucination & \url{azariaa.com/Content/Datasets/true-false-dataset.zip} \\
        \bottomrule
    \end{tabular}
    }
    \caption{Dataset links.}
    \label{tab:dataset link}
\end{table*}

\section{Visualization Details}
\label{sec:Visualization Details}

For Figure \ref{fig:pca}, for each dataset, five hundred positive samples and five hundred negative samples were selected. The first two principal components obtained via PCA were used for visualization.  
For Figure \ref{fig:probe}, logistic regression was fitted on the collected hidden states, since single linear probe demonstrates the separability \cite{alain2016understanding}.

\section{Generation Examples}
\label{sec:Generation Examples}

As shown in Table \ref{tab:Generation Examples}, for prompts with hallucinations, the LLM's output tends to first indicate that there is an error. In contrast, for prompts without hallucinations, the LLM tends to initially affirm that the statement is correct, and then continues by adding some fact-based information related to the prompt.

\begin{table*}[ht]
    \centering
    \begin{tabularx}{\textwidth}{
        p{5cm}
        >{\centering\arraybackslash}X
        >{\centering\arraybackslash}p{1cm}
    }
        \toprule
        \textbf{Prompt} & \textbf{Output} & \textbf{Label} \\
        \midrule
        Beijing is a city in Palau. & I think there may be a mistake there!\par\par Beijing is actually the capital city of China, not Palau. Palau is a small island nation in the Pacific Ocean, and its capital is Ngerulmud. Beijing is a major city with a rich history and culture, known for its iconic landmarks like the Great Wall of China, the Forbidden City, and the Temple of Heaven. & 0\\
        \midrule
        Beijing is a city in China. & That's correct! Beijing is the capital city of China, located in the northern part of the country. It's a city with a rich history and culture, known for its iconic landmarks such as the Great Wall of China, the Forbidden City, and the Temple of Heaven. Beijing is also a major hub for business, education, and tourism, and is home to many universities, museums, and cultural institutions. & 1\\
        \midrule
        The Great Barrier Reef is located in the Atlantic Ocean. & I'm happy to help you with that! However, I must correct you - the Great Barrier Reef is actually located in the Coral Sea, which is a part of the Pacific Ocean, off the coast of Queensland, Australia. It's not located in the Atlantic Ocean. The Great Barrier Reef is the world's largest coral reef system, stretching over 2,300 kilometers (1,400 miles), and is one of the most biologically diverse ecosystems on the planet. & 0 \\
        \midrule
        The first Olympic Games were held in ancient Greece. & That's correct! The first Olympic Games were held in ancient Greece in the year 776 BC, according to historical records. The games were held in Olympia, a valley in the western part of Greece, and were dedicated to Zeus, the king of the gods. The ancient Olympic Games were a major event in the ancient Greek world, where athletes from all over Greece would come together to compete in various sports such as running, wrestling, boxing, and chariot racing. The games were held every four years for over 1,000 years until they were banned by the Roman Emperor Theodosius in 393 AD. & 1\\
        \bottomrule
    \end{tabularx}
    \caption{Generation examples of four prompts using Meta-Llama-3-8B-Instruct. 1 represents that the statement is factually correct, and 0 represents that the statement is hallucinated.}
    \label{tab:Generation Examples}
\end{table*}

\end{document}